\pdfoutput=1

\documentclass[11pt]{article}
\usepackage[table,dvipsnames]{xcolor}

\usepackage[]{acl}

\usepackage{times}
\usepackage{latexsym}

\usepackage{multirow}
\usepackage{multicol}
\usepackage{booktabs}

\usepackage[T1]{fontenc}

\usepackage[utf8]{inputenc}

\usepackage{microtype}

\usepackage{inconsolata}

\usepackage{graphicx}

\title{On the Influence of Context Size and Model Choice in \\ Retrieval-Augmented Generation Systems}

\author{Juraj Vladika \and Florian Matthes \\
  Technical University of Munich \\
  School of Computation, Information and Technology \\
  Department of Computer Science \\
  Garching, Germany \\
  \texttt{\{juraj.vladika, matthes\}@tum.de}}

\begin{document}
\maketitle
\begin{abstract}

Retrieval-augmented generation (RAG) has emerged as an approach to augment large language models (LLMs) by reducing their reliance on static knowledge and improving answer factuality. RAG retrieves relevant context snippets and generates an answer based on them. Despite its increasing industrial adoption, systematic exploration of RAG components is lacking, particularly regarding the ideal size of provided context, and the choice of base LLM and retrieval method. To help guide development of robust RAG systems, we evaluate various context sizes, BM25 and semantic search as retrievers, and eight base LLMs.
Moving away from the usual RAG evaluation with short answers, we explore the more challenging long-form question answering in two domains, where a good answer has to utilize the entire context.
Our findings indicate that final QA performance improves steadily with up to 15 snippets but stagnates or declines beyond that. Finally, we show that different general-purpose LLMs excel in the biomedical domain than the encyclopedic one, and that open-domain evidence retrieval in large corpora is challenging. 
\end{abstract}

\section{Introduction}


The field of Natural Language Processing (NLP) has been vividly transformed with the advent of large language models (LLMs), massive models that excel on a wide range of complex tasks, including text generation, question answering, and summarization \cite{zhao2023survey}. Despite their impressive performance, LLMs have certain limitations. The static nature of the knowledge encoded within their weights can lead to providing outdated content as new information emerges \cite{Zhang2023HowDL}. Furthermore, LLMs can generate plausible sounding but factually incorrect responses (\textit{hallucinations}), as they lack a reliable mechanism to verify the accuracy of the information they produce \cite{10.1145/3571730}. Finally, they can lack specialized knowledge related to advanced expert domains. 

\begin{figure}
\vspace{0.5cm}
    \centering
    \includegraphics[width=0.99\linewidth]{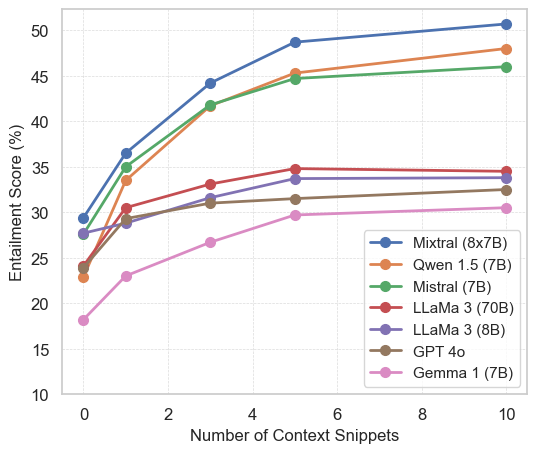}
    \caption{The influence of the number of context snippets passed to the RAG system on the final performance (entailment score) on a biomedical task BioASQ-QA. The performance improves steadily for all models, to a differing extent, and then stagnates after saturation.}
    \label{fig:bioasq-gold}
\end{figure}




To address these shortcomings, the concept of Retrieval-Augmented Generation (RAG) has shown great potential \cite{lewis2020retrieval}. RAG systems enhance the capabilities of LLMs by integrating a retrieval component that allows the model to dynamically utilize external knowledge sources during the generation process. By retrieving relevant information from a curated corpus or the web in real-time, RAG models can produce more accurate, up-to-date, and contextually appropriate responses \cite{fan2024survey}. 
RAG systems have also seen wide adoption in various industry branches, where companies leverage them to build tools for accessing their internal documentation via questions posed in human language \cite{xu2024generative}. 

Despite their increasing popularity and use, there here have been few studies that systematically explore different settings of RAG systems, including the size of the provided context, choice of base LLM, and choice of retriever technique (sparse or dense). While recent work has shown that essential information in long context blocks can get "lost in the middle" \cite{liu-etal-2024-lost} or be affected by noisy context \cite{cuconasu2024power}, most of these studies work with short, factoid question answering (with questions like "\textit{Who won the 2024 Nobel Peace Prize?}") and assume there is one gold context snippet relevant for the answer. There has been less research on how do LLMs use the context for long-form QA, where a holistic final answer has to include multiple or even all context snippets.


To bridge this research gap, our study aims to explore and evaluate various configurations of RAG systems. We systematically investigate how different context sizes, retrieval strategies, and base LLMs impact the performance of RAG systems. We evaluate these parameters through the prism of the generative question-answering task in two different domains: the biomedical BioASQ-QA task \cite{Krithara2023} and the encyclopedic QuoteSum dataset \cite{schuster-etal-2024-semqa}. Both datasets provided the essential resources for our study -- inclusion of \textit{gold evidence snippets} and \textit{human-written answers} utilizing these snippets to answer the questions. By conducting a series of experiments and analyses, we aimed to 
identify best practices for the implementation of RAG systems. 

Our contributions include:
\begin{itemize}
    \item We examine the influence of the number of context snippets in the prompt on the final task performance of the RAG system.
    We observe the performance to steadily improve from 1 to about 10-15 snippets, but then stagnate or even decline by 20-30 snippets.
     \item We test the performance of different LLMs of various sizes in their ability to utilize the context snippets for generating accurate answers.
     The results show Mistral and Qwen to perfrom the best on the biomedical task, while GPT and Llama excel on the encyclopedic task.
    \item We test the open-domain setting, where gold evidence is not known and has to be retrieved from large knowledge bases. We evaluate two different retrievers and show the impact on final performance. We show that this setting is very challenging and performance is far from the gold setting, with the BM25 optimizing for precision, while semantic search gives a wider coverage of retrieved information.
\end{itemize}

We make our code available in a public repository on GitHub.\footnote{\url{https://github.com/jvladika/ContextRAG}}

\section{Related Work}
\subsection{Retrieval-Augmented Generation}
Early approaches to RAG involved simple retrieval and were developed for the task of question answering \cite{chen-etal-2017-reading}. Recent advancements have seen more sophisticated integration of retrieval and generation processes, thereby significantly enhancing the quality and relevance of the generated text \cite{lewis2020retrieval}.
These advancements have been facilitated by improvements in both the retrieval mechanisms, which have become more efficient and effective at finding relevant information, and the generative models, which have become better at integrating and contextualizing the retrieved information \cite{cai2022recent}.

A recent survey by \citet{gao2024retrievalaugmented} separates RAG approaches into \textit{naive RAG} and \textit{advanced RAG}. The naive RAG approach follows a traditional process that includes indexing, retrieval, and generation, also called a “Retrieve-then-Read” framework \cite{zhu2021retrieving}. 
On the other hand, advanced RAG introduces specific improvements to enhance the retrieval quality by employing pre-retrieval and post-retrieval strategies. Pre-retrieval strategies include query rewriting with an LLM \cite{ma-etal-2023-query} or query expansion methods like HyDE \cite{gao-etal-2023-precise}.
Post-retrieval methods focus on selecting essential information from retrieved documents. This includes reranking the retrieved documents with neural models \cite{glass-etal-2022-re2g} or summarizing the retrieved documents before passing them as context \cite{an2021retrievalsum}. 


\subsection{Context and Noise in RAG Systems}

A lot of recent work has explored how to improve RAG and make it more accurate and robust to imperfect context. This includes fact verification \cite{li-etal-2024-llatrieval}, self-reflection with critique \cite{asai2024selfrag}, learning to re-rank the context \cite{yu2024rankrag}, improved answer attribution \cite{kdir24vladika}, adaptive search strategy \cite{jeong-etal-2024-adaptive}, and relevance modeling \cite{wang-etal-2024-rear}. 

There have also been studies exploring the size of input context and its influence on the performance of RAG systems. \citet{liu-etal-2024-lost} highlight the effect of information being \textit{lost in the middle}, showing how RAG mostly focuses on the beginning and the ending of the provided context. Similarly, \citet{cuconasu2024power} examine the influence of the position of the most relevant snippet in the context and the influence of noisy snippets on the performance. Both of these studies work with factoid QA dataset where it is assumed one context snippet is the most important for the answer.

\citet{xiong-etal-2024-benchmarking} analyze the effect of number of context snippets on five multiple-choice biomedical QA tasks, while \citet{vladika-matthes-2024-improving} analyze the impact of the number of snippets as well as context recency and popularity for biomedical QA. \citet{10.1609/aaai.v38i16.29728} evaluated the noise robustness and context integration of different LLMs for RAG. Most similar to our work is the study by \citet{hsia2024ragged}, where the influence of different RAG components is tested with eight LLMs and it also includes BioASQ as a benchmark dataset.

While these studies have discovered important principles in context inclusion for RAG systems, they predominantly evaluate it on multiple-choice or short-form QA tasks where there is one clear answer and one most important context snippet. Our work evaluates generative question answering where potentially all snippets could be relevant for inclusion in the answer, which is a more challenging setting. Additionally, we provide a comprehensive evaluation of three main RAG components: the influence of the context size, different retrieval techniques, and choice of base LLMs.


\section{Foundations}
\subsection{RAG System for Question Answering}

Typically, a RAG system consists of a \textit{retriever} and a \textit{reader}. Retriever has to search and collect relevant evidence snippets that are passed as \textit{context} inside of a prompt to the reader. Our study investigates the importance of those three aspects (retriever, context, reader) on the final performance of the whole system. We first focus on the influence of context size on the readers' QA capability, followed by the importance of choosing the reader by comparing different base LLMs on the task, and finally, we test the influence of two different retrievers on the final QA performance (BM25 and semantic search). To formally define: Given a question $q$ and context $c$ consisting of context snippets $c_1, c_2, ..., c_n$, the goal is to generate an answer $a$ with a model \textit{reader(q, c) = a}. The context $c$ is provided in the first experiment, but in an open-domain setting, given a document corpus $D$ with documents $d_1, d_", ... d_N$, the idea is for a retriever to \textit{retrieve(q, D) = d1, d2} best matching documents and then from them extract context snippets. 

\subsection{Datasets}
\noindent \textbf{BioASQ-QA} \cite{Krithara2023} is a biomedical question answering (QA) benchmark dataset in English. 
It has been designed to reflect real information needs of biomedical experts. 
The questions are written by biomedical experts and the evidence corpus used to answer them is PubMed \cite{white2020pubmed}, the large database of biomedical research papers.
The dataset is a part of the ongoing shared challenge, and we use the 2023 version, Task 10b. While the full dataset contains various types of questions (yes/no, factoid, lists), we utilize only the so-called \textit{summary} questions -- questions paired with human-selected evidence snippets from PubMed abstracts and human-written "ideal answers", which are essentially natural language summaries of the provided snippets. In total, there are 1130 summary questions.

\noindent \textbf{QuoteSum} \cite{schuster-etal-2024-semqa} is a dataset of encyclopedic questions, relevant passages, and human-written semi-extractive answers. The questions are human-written and are paired with up to 8 passages (evidence snippets) from Wikipedia. These passages are used as the main source by annotators to write the answers. QuoteSum contains 805 instances and covers various domains such as geography, history, arts, and technology. An example question is "\textit{Why was Stonehenge built in the first place?}".

These datasets contain the gold evidence snippets and human-written answers based on the snippets, making them a suitable testbed for our study. While BioASQ might be difficult given its language rich with complex biomedical terminology, the main challenge is in successfully utilizing the given context and summarizing it into a concise but informative answer. We intentionally do not benchmark on any biomedical LLM to not give any model a possible advantage.


\begin{table*}[htpb]
\small
\centering
\rowcolors{2}{gray!25}{white}

\begin{tabular}{r|ccc|ccc|ccc|ccc}

\rowcolor{gray!40}
\textbf{\# docs} & \multicolumn{3}{c}{\textbf{GPT 3.5}} & \multicolumn{3}{c}{\textbf{GPT 4o}} & 
\multicolumn{3}{c}{\textbf{LLaMa 3 (70B)}} & \multicolumn{3}{c}{\textbf{Mixtral (8x7B)}}  \\
\cline{1-13}
\rowcolor{gray!40}
  \textbf{}    & \textbf{R-L}       & \textbf{BSc}       & \textbf{Ent.\%}  & \textbf{R-L}       & \textbf{BSc}       & \textbf{Ent.\%}       & \textbf{R-L}       & \textbf{BSc}       & \textbf{Ent.\%}        & \textbf{R-L}       & \textbf{BSc}       & \textbf{Ent.\%}       \\
\hline
 & \multicolumn{12}{c}{\textsc{BioASQ}} \\

\textbf{0}       &   23.2     &    87.1     &   22.5    &   23.8  &   87.0   &   23.9    &   22.9     &  86.9    &  24.1   &    21.8     &   85,8    &   29.4         \\ \hline
\textbf{1}       &   28.0     &    87,9     &   29.3    &   28.2  &   87.9   &   29.3     &    28.3    &   87,8   &  30.5  &    29.6     &    87.9   &   36.5         \\
\textbf{3}       &   30.9    &      88.5   &   31.4   &   31.1    &   88.4    &   31.0  &    31.4    &   88.4   &  33.1  &    34.8   &   88.9   &   44.2          \\
\textbf{5}     &      31.9     &    88.6   &    32.0     &     31.9  &  88.6   &     31.5     &    32.0  &  88.5  &  \textbf{34.8}   &    36.4    &    89.0  &    48.7           \\
\textbf{10}      &   \textbf{32.7}    &   \textbf{88.8  }    &   \textbf{32.6}    &    \textbf{32.8}    &   \textbf{88.8}   &    \textbf{32.5}    &  \textbf{32.2}     &  \textbf{88.6}  &   34.5  &    \textbf{37.7}    &   \textbf{89.2}    &   \textbf{50.7}         \\ \hline

& \multicolumn{12}{c}{\textsc{QuoteSum}} \\

\textbf{0} & 27.2 & 85.0 & 20.4 & 26.9 & 84.9 & 20.3 & 26.3 & 84.4 & 21.8 & 22.3 & 83.6 & 15.2 \\
\textbf{1} & 36.4 & 87.1 & 41.7 & 36.6 & 87.1 & 41.7 & 34.3 & 86.5 & 36.5 & 33.8 & 86.1 & 32.0  \\
\textbf{3} & 39.0 & 87.6 & 42.9 & 39.1 & 87.7 & 43.1 & 37.4 & 87.5 & 40.5 & 37.2 & 86.8 & 35.2  \\
\textbf{5} & \textbf{39.9} & 87.7 & 44.0 & \textbf{39.7} & 87.7 & \textbf{44.2} & 38.4 & 87.5 & \textbf{41.9} & 37.4 & 86.9 & \textbf{36.1}  \\
\textbf{10} & 39.7 & \textbf{87.7} & \textbf{44.2} & 39.6 & \textbf{87.7} & 43.4 & \textbf{39.4} & \textbf{87.5} & 41.8 & \textbf{37.7 } & \textbf{86.9} & 35.9  \\
\end{tabular}

\caption{\label{tab:results_gold_big} Results of final QA performance on BioASQ and QuoteSum for different number of \textbf{gold snippets}, using the \textbf{four big LLM}s as readers: GPT 3.5, GPT 4o, Mixtral (8x7B), LLaMa 3 (70B). The results are measured with ROUGE-L (R-L), BERTScore (BSc), and average entailment prediction of the NLI model (Ent.\%).}

\end{table*}

\section{Experiment}

\subsection{Models}
We conduct our experiments using a multitude of different LLMs that serve as readers, i.e., the models reading and comprehending the context and then generating the answers from it. The experiments were mainly conducted in June 2024 and reflect the up-to-date LLM landscape of that time.

We start with \textbf{GPT} as a commercial state-of-the-art LLM in our comparison since it has demonstrated remarkable zero-shot performance on various NLP tasks. Consequently, it is often used as a benchmark for comparing LLMs. We use the \textbf{GPT-3.5} (Turbo-0125) as the standard ChatGPT version, and also \textbf{GPT-4o} (Turbo-0513), the updated "omni" version of GPT-4 \cite{Achiam2023GPT4TR}, which was shown to improve the performance. We then also include two popular open-weights models that achieved impressive performance, namely \textbf{Mixtral} (8x7B) \cite{jiang2024mixtral}, based on a sparse mixture-of-experts architecture \cite{fedus2022review}; and \textbf{LLaMa 3} (70B) \cite{llama3modelcard}, a powerful staple model from Meta. All models are instruction-tuned ("chat") versions. 

For the smaller language models, we choose \textbf{Mistral-7B} (Instruct-v3) \cite{jiang2023mistral}, the smaller counterpart to Mixtral; then the \textbf{Gemma} (1) \cite{Mesnard2024GemmaOM}, a lightweight open model from Google built from the research and technology used to create Gemini models \cite{geminiteam2024gemini}; and the smaller, 8B version of \textbf{LLaMa 3}. We additionally benchmark \textbf{Qwen 1.5 (7B)} (Chat), another recently popular and powerful language model \cite{bai2023qwen}. All of these models are open-source models, and we use the instruction-tuned versions. 

\subsection{Setup}
We use the same prompt and setup for all of the benchmarked models:

\fbox{\begin{minipage}{16em}
Give a simple answer to the question based on the context.

QUESTION: 

<the current question>

CONTEXT: 

[snippet$_1$, snippet$_2$, ..., snippet$_n$] 

ANSWER:
\end{minipage}}

For the internal-knowledge setting with no context, the instruction was changed accordingly to \textit{Give a simple answer to the question based on your best knowledge.} and the \textit{CONTEXT} part removed. While it would have been an interesting experiment to also give the LLMs few-shot examples of QA pairs, we intentionally opt for this zero-shot setting so that the focus of the experiments lies solely on the utilization of provided context for answering and not on potential in-context learning abilities. 

GPT models were prompted through the OpenAI API, while all of the open-source models were queried with API calls through the Together AI service\footnote{\href{https://docs.together.ai/reference/chat-completions}{Together AI: https://docs.together.ai/}} platform, which hosts many popular open-source models. We set the token limit to 512 and the temperature parameter to 0, maximizing deterministic generation by favoring high-probability words and thus ensuring reproducibility of the results. One run through the whole dataset with five settings took two computation hours. For embedding models, we used one NVidia V100 GPU card with 16 GB of VRAM.

\begin{table*}[htpb]
\small
\centering
\rowcolors{2}{gray!25}{white}

\begin{tabular}{r|ccc|ccc|ccc|ccc}

\rowcolor{gray!40}
\textbf{\# docs} & \multicolumn{3}{c}{\textbf{Gemma (7B)}} & 
\multicolumn{3}{c}{\textbf{LLaMa 3 (8B)}} & \multicolumn{3}{c}{\textbf{Mistral (7B)}} & \multicolumn{3}{c}{\textbf{Qwen (7B)}}  \\
\cline{1-13}
\rowcolor{gray!40}
  \textbf{}    & \textbf{R-L}       & \textbf{BSc}       & \textbf{Ent.\%}  & \textbf{R-L}       & \textbf{BSc}       & \textbf{Ent.\%}       & \textbf{R-L}       & \textbf{BSc}       & \textbf{Ent.\%}        & \textbf{R-L}       & \textbf{BSc}       & \textbf{Ent.\%}       \\
\hline
 & \multicolumn{12}{c}{\textsc{BioASQ}} \\
 
\textbf{0}         &   20.7  &   79.7   &   18.2    &   21.1     &  85.2    &  27.7   &    20.6     &   85.4    &   27.6  & 20.3 & 85.5 & 22.9      \\ \hline
\textbf{1}       &   25.9  &   85.7   &   23.0     &    28.5    &   87.8   &  28.8  &    28.7     &    87.9   &   35.0    &    28.5     &    87.7   &   33.5       \\
\textbf{3}       &   30.6  &   86.7   &   26.7       &    32.2   &   88.2  &  31.6      &    33.1    &   88.7   &   41.8   &    33.4     &    88.9   &   41.7        \\
\textbf{5}       &     32.2 &  87.0   &     29.7     &    36.7  &  \textbf{88.5 } &  33.7   &    34.7    &    88.9  &    44.7   &    35.2     &    89.0   &   45.3         \\
\textbf{10}     &    \textbf{33.5 } &   \textbf{87.2}   &   \textbf{ 30.5}    &  \textbf{37.3 }    &  88.4  &   \textbf{33.8}  &   \textbf{ 36.4}    &   \textbf{89.2  }  &   \textbf{46.0}   &    \textbf{36.8 }    &   \textbf{ 89.1 } &   \textbf{48.0  }     \\  \hline

 & \multicolumn{12}{c}{\textsc{QuoteSum}} \\

\textbf{0} & 8.7 & 67.4 & 9.8 & 24.0 & 83.6 & 18.5 & 25.2 & 83.9 & 13.6 & 25.2 & 84.3 & 17.3 \\
\textbf{1} & 15.8 & 54.6 & 13.7 & 33.8 & 86.4 & 37.2 & 34.5 & 86.6 & 35.4 & 35.7 & 86.9 & 39.6 \\
\textbf{3} & 25.2 & \textbf{77.1} & 22.0 & 38.8 & 87.0 & 38.3 & 37.4 & 87.2 & 38.0 & 39.0 & 87.4 & 42.2 \\
\textbf{5} & \textbf{ 24.8} & 76.4 & 22.4 & 39.4 & 87.1 & 37.7 & 38.6 & \textbf{87.3} & 39.3 & 39.3 & 87.5 & \textbf{43.0} \\
\textbf{10} & 24.7 & 76.1 & \textbf{ 22.4 } & \textbf{39.7} & \textbf{87.1} & \textbf{38.0} & \textbf{39.0} & 87.2 & \textbf{39.5} & \textbf{39.3} & \textbf{87.5} & 42.9 \\

\end{tabular}

\caption{\label{tab:results_gold_small} Results of final QA performance on BioASQ and QuoteSum for different number of\textbf{ gold snippets}, using the \textbf{four small LLMs } as readers: Gemma (7B), LLaMa 3 (8B), Mistral v3, and Qwen 1.5 (7B). The results are measured with ROUGE-L (R-L), BERTScore (BSc), and average entailment prediction of the NLI model (Ent.\%).}

\end{table*}

\subsection{Experiment Rounds}

\paragraph{Context Size and Reader Performance.} The first round of experiments consisted of varying the number of context snippets passed in the prompt and observing how the QA performance changes. We pass the gold snippets in this experiment since they are known to us and the focus is only on the quantity (context size). While it seems intuitive that adding more snippets will improve the final scores, because an answer based on partial information will be incomplete, we wanted to test: (1) to which extent do different LLMs utilize the provided context, and (2) when do the LLMs get saturated with too much context, leading to stagnation or decline.

As a starting point, we first pose the question to LLMs with no context (0 snippets), thus testing their \textit{internal knowledge recall}. While an interesting research caveat on its own, we use it here just as a baseline. Afterward, we vary the numbers of context snippets in the array of 1, 3, 5, 10; to give the idea of a general trend. In case a question has fewer snippets than the given \textit{k}, then all of the snippets for that question were used. In BioASQ, more than 80\% of all questions have at least 3 snippets, around 60\% at least 5 snippets, and around one third at least 10 snippets -- using more than 10 wouldn't make a lot of sense given that only 18\% of questions have more than 10 snippets. Similarly, QuoteSum has around 75\% questions with at least 3 snippets, 50\% with at least 5, and 30\% have the maximum 8 snippets (labeled "10" in tables for consistency).

\paragraph{Closed Retrieval} Apart from the easier setting where gold snippets are provided to the model, we also explore the more challenging setup with evidence retrieval "in the wild" \cite{chen-etal-2024-complex}. In this case, the RAG system first has to retrieve the evidence from a knowledge base before producing an answer based on it. For the closed retrieval setting in BioASQ, we only use the PubMed documents required to answer its 1130 questions. This results in around 8000 documents as a knowledge base. This mimics the common RAG use case in the industry where one would be working with a limited knowledge corpus containing internal company documents. The abstracts are saved in a vector database and embedded with a sentence embedding model (we use the biomedical \textit{S-PubMedBERT-MS-MARCO}\footnote{\url{https://huggingface.co/pritamdeka/S-PubMedBert-MS-MARCO}} from \citealp{deka2022improved}). Afterward, the top 10 most similar documents to the question are retrieved (using cosine similarity and the embedding model), and the most similar sentence from each document is selected as an evidence snippet. The amount of selected sentences/documents is also varied with amounts 1, 3, 5, 10; to align with the first round of experiments. For QuoteSum, we omit this experiment as the subset of Wikipedia articles is not provided.

\begin{table*}[htpb]
	\small
	\centering
	\rowcolors{2}{gray!25}{white}
		
		\begin{tabular}{r|ccc|ccc|ccc|ccc}
			
			\rowcolor{gray!40}
			\textbf{\# docs} & \multicolumn{3}{c}{\textbf{GPT 3.5}} & \multicolumn{3}{c}{\textbf{GPT 4o}} & 
			\multicolumn{3}{c}{\textbf{LLaMa 3 (70B)}} & \multicolumn{3}{c}{\textbf{Mixtral (8x7B)}}  \\
			\cline{1-13}
			\rowcolor{gray!40}
			\textbf{}    & \textbf{R-L}       & \textbf{BSc}       & \textbf{Ent.\%}  & \textbf{R-L}       & \textbf{BSc}       & \textbf{Ent.\%}       & \textbf{R-L}       & \textbf{BSc}       & \textbf{Ent.\%}        & \textbf{R-L}       & \textbf{BSc}       & \textbf{Ent.\%}       \\
			\hline
			\textbf{0}       &   23.2     &    87.1     &   22.5    &   23.8  &   87.0   &   23.9    &   22.9     &  86.9    &  24.1   &    21.8     &   85.8    &   29.4         \\ \hline
			\textbf{1}       &   24.2     &    87.8     &   25.2    &   24.5  &   87.2   &   25.4    &    24.0    &   86.9   &  25.8  &    25.7     &    87.0   &   30.1         \\
			\textbf{3}       &   32.7    &      88.0   &   27.4   &   34.5  &   87.9   &   27.2       &    28.1    &   87.7   &  27.7      &    30.3   &   87.8   &   36.1         \\
			\textbf{5}     &      30.5    &    88.1   &    28.6    &     30.0 &  88.2  &     28.8     &    29.3  &  88.0  &  30.0   &    31.6    &    87.9  &    39.9          \\
			\textbf{10}      &   \textbf{32.0 }   &   \textbf{88.7}     &   \textbf{31.4}    &    \textbf{31.8}  &   \textbf{88.6}   &    \textbf{31.0 }   &  \textbf{31.0 }    &  \textbf{88.3}  &   \textbf{32.1}  &    \textbf{32.9}   &   \textbf{88.2}    &   \textbf{44.4}         \\
			
		\end{tabular}
	
	\caption{\label{tab:results_retriv_snippets} Results of final QA performance on BioASQ for different number of \textbf{retrieved context} snippets in the \textbf{closed retrieval} setting (with a corpus of 8 thousand PubMed documents), using the four big LLMs. The results are measured with ROUGE-L (R-L), BERTScore (BSc), and average entailment prediction of the NLI model (Ent.\%).}
	
\end{table*}

\paragraph{Open Retrieval.} Finally, we test the QA performance of both datasets in the most challenging open setting -- using a large knowledge base where the retriever first has to sift through millions of documents to discover the most relevant ones. For BioASQ, we test two different retrievers -- semantic search with the same sentence embedding model as in last round (\textit{S-PubMedBERT-MS-MARCO}) and the sparse retrieval technique BM25, which has a long-established track record of good performance for information retrieval (IR) tasks. 

For BioASQ, we use MEDLINE, a snapshot of currently available abstracts in PubMed that is updated once a year. We used the 2022 version found on the official website.\footnote{\url{https://www.nlm.nih.gov/databases/download/pubmed_medline.html}} We filter it to a 10-year span from 2012 to 2022, following BioASQ's time range -- this results with 10.6 million abstracts in total. For QuoteSum, we use Wikipedia since the dataset is based on it. We query the Wikipedia search API directly through a link.\footnote{\url{https://en.wikipedia.org/w/api.php?action=query&list=search&srsearch=\{text\}&format=json}} Wikipedia search is based on BM25.\footnote{Source: \url{https://wikimedia-research.github.io/Discovery-Search-Test-BM25/}} While a popular way to benchmark retrievers is using common IR metrics like \textit{recall@k}; we focus only on benchmarking the final QA performance, as this both keeps it consistent with the previous experiments and also highlights the fact that the final answer is the most important artefact of a QA system.

Unlike gold snippets where we often only had up to 10 provided in the original dataset, the open setting allows to keep increasing the number of snippets indefinitely. Therefore, we additionally evaluate with 15, 20, 30 snippets, to test the effect of context saturation.

\subsection{Evaluation}
To evaluate the quality of the generated answers, we use three main metrics. Given that the dataset contains ideal answers, we can use reference-based metrics. Evaluating LLMs for long-form QA is a challenging, ongoing research problem, and no metric is ideal \cite{xu-etal-2023-critical}. Still, we cover a variety of metrics, to gain an overview.

The first metric is  \textbf{ROUGE} \cite{lin-2004-rouge},\footnote{\url{https://pypi.org/project/rouge/}} which looks at the recall between the reference answer and the generated answer. Specifically, we use the ROUGE-L, which looks at the longest overlapping sequence between the reference and generated answer. Since this metric focuses solely on lexical overlaps, we use two additional semantic metrics.  We also apply the \textbf{BERTScore} metric, which captures semantic similarity by using the BERT model's embeddings \cite{zhang2020BERTScore}.

The third metric utilizes the concept of natural language inference (\textbf{NLI}),
by using the reference answer as the hypothesis and the generated answer as the premise. The intuition behind this approach is that a good answer should logically entail the reference answer. Using NLI this way has been done to evaluate the quality of summaries and text generation \cite{laban-etal-2022-summac}.
We use the model DeBERTa-v3 \cite{he2023debertav}, which was shown to work well with NLI and reasoning tasks 
We use the version \textit{Tasksource} that was fine-tuned on a wide array of NLI datasets and other classification datasets \cite{sileo2023tasksource}.\footnote{\url{https://huggingface.co/sileod/deberta-v3-base-tasksource-nli}} This model predicts three scores (entailment, neutral, contradiction) and we report on the average \textbf{entailment} score as \textbf{Ent\%}.

We additionally use for the first experiment \textbf{METEOR} \cite{lavie-agarwal-2007-meteor} (in \textit{nltk}), a metric that looks at word overlaps like ROUGE but relaxes the matching criteria -- it takes into account word stems and synonyms. 
Finally, we also report on the average \textbf{cosine similarity} (Cos) of text embeddings between generated and reference answers, a metric that emphasizes the semantic similarity of these two strings. For that we use the sentence transformer \textit{all-mpnet-base-v2}.\footnote{\url{https://huggingface.co/sentence-transformers/all-mpnet-base-v2}} The results are in Tables \ref{tab:results_metcos_bioasq} and \ref{tab:results_metcos_semqa} in Appendix.

\begin{table*}[htpb]
\small
\centering
\rowcolors{2}{gray!25}{white}


\begin{tabular}{r|cccccc|cccccc}

\rowcolor{gray!40}
\textbf{\# docs} & \multicolumn{3}{c}{\textbf{GPT 4o (semantic)}} & \multicolumn{3}{c}{\textbf{Mixtral (semantic)}} & 
\multicolumn{3}{c}{\textbf{GPT 4o (BM25)}} & \multicolumn{3}{c}{\textbf{Mixtral (BM25)}}  \\
\cline{1-13}
\rowcolor{gray!40}
  \textbf{}    & \textbf{R-L}       & \textbf{BSc}       & \textbf{Ent.\%}  & \textbf{R-L}       & \textbf{BSc}       & \textbf{Ent.\%}       & \textbf{R-L}       & \textbf{BSc}       & \textbf{Ent.\%}        & \textbf{R-L}       & \textbf{BSc}       & \textbf{Ent.\%}       \\
\hline

\textbf{0}       &    23.8  &   87.0   &   23.9    &     21.8     &   85.8    &   29.4      &    23.8  &   87.0   &   23.9  &    21.8     &   85.8    &   29.4             \\ \hline
\textbf{1} & 20.9 & 86.5 & 18.9 & 22.1 & 86.0 & 23.4 & 20.7 & 86.5 & 18.9 & 22.2 & 86.0 & 23.5 \\
\textbf{3} & 23.8 & 86.6 & 19.0 & 23.1 & 86.1 & 23.9 & 22.4 & 86.7 & 20.1 & 22.8 & 86.0 & 24.1 \\
\textbf{5} & 22.9 & 86.7 & 19.5 & 23.3 & 86.0 & 25.8 & 23.2 & 86.9 & 20.9 & 23.0 & 86.1 & 26.1 \\
\textbf{10} & 23.0 & 86.8 & 19.9 & 23.2 & 86.0 & 27.6 & 23.4 & 86.9 & 21.5 & 23.3 & 86.0 & 28.9 \\ \hline
\textbf{15} & 25.1 & 87.1 & 26.9 & 24.7 & 86.2 & 31.4  & 24.9 & 87.2 & 26.5 & 25.0 & 86.4 & 31.1 \\
\textbf{20} & 25.3 & \textbf{87.3} & \textbf{27.6} & 24.6 & 86.2 & \textbf{31.9} & \textbf{25.5} & \textbf{87.4} & \textbf{27.9} & \textbf{25.2} & \textbf{86.5} & \textbf{32.0}  \\
\textbf{30} & \textbf{25.7} & 87.2 & 27.5 & \textbf{24.9} & \textbf{86.3 } & 31.5 & 25.4 & 87.2 & 27.4 & 25.1 & 86.3 & 31.6  \\

\hline
  
\end{tabular}

\caption{\label{tab:results_retriv_docs} Results of final QA performance on BioASQ for different number of \textbf{retrieved context} snippets in the \textbf{open retrieval} setting, using PubMed (10 million doc.) with two big LLMs. Semantic refers to semantic search using dense vector embeddings and BM25 is a sparse retrieval technique, which showed better performance here. The results are measured with ROUGE-L (R-L), BERTScore (BSc), and average entailment prediction of the NLI model (Ent.\%).}

\end{table*}

\begin{table}[htpb]
\small
\centering
\rowcolors{2}{gray!25}{white}

\begin{tabular}{r|ccc|ccc}

\rowcolor{gray!40}
\textbf{\# } & \multicolumn{3}{c}{\textbf{GPT 4o}} & \multicolumn{3}{c}{\textbf{Mixtral}}  \\

\rowcolor{gray!40}
  \textbf{}    & \textbf{R-L}       & \textbf{BSc}       & \textbf{E.\%}  & \textbf{R-L}       & \textbf{BSc}       & \textbf{E.\%}     \\
\hline
\textbf{1} & 23.0 & 83.7 & 12.9 & 25.4 & 84.7 & 16.7 \\
\textbf{3} & 24.3 & 84.0 & 13.6 & 26.2 & 84.9 & 16.4 \\
\textbf{5} & 24.8 & 84.1 & 14.4 & 26.4 & 84.9 & 17.2 \\
\textbf{10} & 25.4 & 84.2 & 15.4 & 27.2 & 85.0 & 17.3 \\
\textbf{20} & 25.8 & \textbf{84.4 }& \textbf{16.3} & 27.8 & \textbf{85.2} & \textbf{18.9} \\
\textbf{30} & \textbf{26.1} & 84.2 & 16.2 & \textbf{28.0} & 85.1 & 18.8 \\
\hline  
\end{tabular}

\caption{\label{tab:results_retriv_quotesum} Results of final QA performance on QuoteSum for different number of \textbf{retrieved context} snippets from \textbf{retrieved documents} from Wikipedia, using Wikipedia's built-in BM25-based search.}

\end{table}

\section{Results}

\subsection{Gold Snippets}
The results of four large LLMs with gold snippets are present in Table \ref{tab:results_gold_big}. All models observe a similar pattern: after starting with a rather low zero-shot performance, already utilizing just one context snippet leads to a big jump in performance. After that, most models slowly and steadily improve their answers as measured by all three metrics. Looking at different models, for BioASQ, GPT 4o and LLaMa 3 (70B) had a similar performance, with LLaMa slightly outperforming GPT. Mixtral showed for BioASQ by far the strongest performance among the models across all three metrics.
The biggest jump is observed in the entailment metric, showing how the answers generated by Mixtral had a higher entailment score -- meaning a higher logical alignment with the reference answer.
On the other hand, for QuoteSum, the situation is the other way around. GPT models performed the best, followed by LLaMa, and Mixtral came in last place. The zero-context performance was a lot lower than any context-based setting, showing how questions from this dataset are highly dependent on context. 

The difference in performance for BioASQ could be explained by the different levels of biomedical knowledge that some models encode compared to others. In related studies, Mixtral and Mistral were found to encode more recent and higher quality biomedical knowledge than GPT 4 \cite{vladika-etal-2024-medreqal}, while Mistral was found to perform better on biomedical QA tasks than the domain-specific biomedical variation BioMistral \cite{dada2024doesbiomedicaltraininglead}. 

The results of four smaller LLMs with gold snippets are shown in Table \ref{tab:results_gold_small}. A similar pattern is observed -- the top-1 snippet helps improve the performance significantly, while further additions lead to a more linear improvement. This holds true for LLaMa 3 (8B) and Gemma (7B). Mistral, just like its larger counterpart Mixtral, led to excellent performance as measured by all three metrics. The best performance for top-5 and top-10 was done by Qwen (7B). For BioASQ, even compared to the way bigger models LLaMa 3 (70B) and GPT 4o in the previous table, Mistral and Qwen demonstrated a lot stronger performance and context utilization, showing that the model size is not always the most important factor in RAG systems and comprehensive answering. For QuoteSum, Qwen outperfromed LLaMa 3 (70B), while Mistral outperformed its larger counterpart Mixtral.

Table \ref{tab:six_answers} shows an example of a question from BioASQ, its top 3 gold snippets, and the ideal answer from our dataset. It then shows the responses generated by different LLMs for the given question and given three snippets, demonstrating the differences in addressing the question based on the given context. Mixtral seems to have picked up the most details while keeping it concise, Qwen gave the longest answer, while GPT 4o and LLaMa 3 (70B) left some crucial details out. Interestingly, LLaMa 3 (8B) gave a more informative answer than its bigger counterpart.

\begin{table*}[htpb]
\small
    \centering
    \begin{tabular}{p{25mm}|p{31mm}p{41mm}p{45mm}}
        \toprule
        \textbf{Question} & \textbf{Gold answer} & \textbf{Mixtral (w/ top 3 snippets)} & \textbf{Mixtral (internal knowledge)} \\ \midrule
        For what is Protein A from Staphylococcus aureus used in biochemistry? & Protein A from the bacterium Staphylococcus aureus (SpA) is used as an \textbf{affinity ligand for purification of immunoglobulin G (IgG)}. & Protein A from Staphylococcus aureus is used in biochemistry as a powerful diagnostic reagent, and it also has applications in human disease therapy. It is isolated from the bacterial pathogen and is known to be produced exclusively as an extracellular protein in certain strains. & Protein A from Staphylococcus aureus is often used in biochemistry and molecular biology for the \textcolor{teal}{purification of antibodies} and the detection of antibody-antigen complexes. It has a \textcolor{teal}{high affinity} for the Fc region of \textcolor{teal}{IgG antibodies}, making it a useful tool in various laboratory techniques such as immunoprecipitation, immunoblotting, and ELISA. \\ \hline 
        
         What is the function of Taraxasterol in rheumatoid \newline arthritis? & Taraxasterol suppresses \textbf{inflammation} in \textbf{rheumatoid arthritis}. &  Based on the context provided, \textcolor{BrickRed}{there is no direct information} about the function of Taraxasterol in rheumatoid arthritis. & Taraxasterol is a plant-derived compound that has been studied for its potential \textcolor{teal}{anti-inflammatory effects}, which are beneficial in managing symptoms of \textcolor{teal}{rheumatoid arthritis}.
         \\ \hline
    \end{tabular}
    \caption{Two examples where internal knowledge of LLMs gave a better answer than the RAG answer based on top 3 retrieved snippets}
    \label{tab:protein_a_swapped}
\end{table*}

\subsection{Closed Retrieval}

This setting used the small knowledge base of eight thousand PubMed articles that were used as gold evidence in BioASQ. The results of the experiments are shown in Table 3. In this setting, it is visible that the performance dropped when compared to Table 1. Even in Mixtral, which was the best performing model, the performance dropped on average. Still, the performance kept improving with each increase of \textit{top k} snippets selected, once again demonstrating that with more context, the performance was better. This was especially apparent in the top-10 setting, since the more evidence snippets selected, the higher the chances of selecting some of the gold evidence snippets used for the generation of the ideal answer.

\subsection{Open Retrieval}

The final setting used around 10 million PubMed articles as its knowledge base for retrieval. The idea of this experiment is to see (1) how much the performance in the open setting differs from the closed setting with gold evidence and (2) what the influence of different retrievers is on this performance. 
Results for BioASQ are shown in Table \ref{tab:results_retriv_docs}, while results for QuoteSum are shown in Table \ref{tab:results_retriv_quotesum}. Since we stored an offline copy of PubMed documents, we could use both BM25 and semantic search (with local vector embeddings), while for Wikipedia, we used its search API based on BM25.

When compared to previous tables, it is evident that open retrieval is the most challenging setting, with lowest average scores overall. It is also interesting to observe that retrieving the documents with BM25 led to a slightly better final performance compared to semantic search.

\section{Discussion}
\subsection{Retrieval Techniques}
Looking at Table \ref{tab:results_retriv_snippets}, BM25 led to a better performance overall. Given that it works with keyword matches, this retrieval technique optimizes for precision in search results rather than recall, thus ensuring that more documents will actually be discussing the same concepts (words) mentioned in the question itself. This shows that optimizing for precision and matching the keywords of the query to the knowledge contained in the knowledge base can lead to improved performance. Especially in critical applications like the biomedical domain of question answering, optimizing for precision and robust answers can be more important than the recall provided by semantic search. 

\subsection{Internal vs. External Knowledge Conflict}
An interesting remark from open retrieval in Table \ref{tab:results_retriv_docs} is that both GPT and Mixtral have better scores for their zero-shot answers (with 0 context snippets) than the answers where up to 10 context snippets were provided. After we analyzed many outputs, a potential explanation of this phenomenon is that, while snippets discovered in the corpus can be semantically similar to the question, they do not always provide all the important information. On the other hand, when using just the vanilla prompt, LLM answers based on its "internal" knowledge –- these answers reflect the collected knowledge of LLMs gained from the large pre-training corpora. Therefore, the internal LLM answers can be more informative than the RAG setting where an LLM is instructed to answer only using the provided short snippets. As more snippets are added, the informativeness of RAG answers starts surpassing LLM's internal knowledge. Recent studies have also observed that for biomedical tasks, it can sometimes be more beneficial to generate internal answers than retrieve external context \cite{frisoni-etal-2024-generate}.

Consider the first example in Table \ref{tab:protein_a_swapped} -- the answer from Mixtral's internal knowledge mentions purification and IgG, the same as in the gold answer, while the answer based on the top 3 snippets produced an incomplete answer. In general, the bottleneck is often tied to incorrect retrieval –- sometimes, the retrieved snippets did not address the question at all, especially for complex biomedical terms found in BioASQ. On the other hand, LLMs in the vanilla setting will always provide the answer based on their best knowledge, thus outperforming cases of bad retrieval. This is apparent in the second example in Table 6. This demonstrates the well-known challenges of knowledge conflict between the internal knowledge of LLMs and the knowledge passed to them in the context \cite{marjanovic-etal-2024-dynamicqa} and is an interesting future research direction following from our study.

\subsection{Context Saturation}
Another insight of the study visible in Table \ref{tab:results_retriv_docs} is that there is a certain upper limit to the performance improvements. As we kept on adding more and more context, increasing to 20, the performance stalled and then slightly dropped for 30 retrieved context snippets. As the saturation point is reached, adding more context to the prompt just leads to noise and confusion in answering. This confirms the previous findings from literature that context can get "lost in the middle" of long prompts and ignored by the reader LLM when answering the questions \cite{liu-etal-2024-lost, hsieh-etal-2024-found}.

\section{Conclusion}
In this study, we explored the effectiveness of Retrieval-Augmented Generation (RAG) systems for long-form question answering using two datasets. We systematically evaluated the impact of various settings of retrieval strategies, context sizes, and base LLMs on RAG performance. Our findings indicate that increasing context snippets enhances performance up to around 15 snippets. For biomedical QA, models like Mixtral and Qwen performed the best, while they were outperformed by GPT-4o and LLaMa 3 for encyclopedic QA. In open retrieval setting, BM25 yielded better results for biomedical QA, with open challenges remaining for exploring knowledge conflict between internal LLM knowledge and external context. We envision future work to explore the effects of query expansion methods and evidence re-ranking.
We hope our work provides valuable insights for optimizing applied RAG systems in practice.

\section*{Limitations}
Our study is limited to two datasets, thus making it possible that some findings would not universally generalize to different domains and tasks. Additionally, we only evaluate the models in a zero-shot setting, whereas a few-shot setting with some examples of questions and answers would have led to a more uniform performance across models. 

The use of automated metrics for natural language generation tasks is not ideal, and they have certain drawbacks. ROUGE score focuses too much on word overlaps with no semantic matching, BERTScore often gives scores in a very tight range, and NLI models can struggle with long text as input. Ideally, a human evaluation would bring a more rigorous result assessment, but hiring human annotators, especially domain experts for the medical text, was outside of our scope and resources. 

Finally, the LLMs, embedding models, and retriever models tested in this study represent only a subset of the quickly evolving landscape of NLP models and technologies. We selected some of the most popular and trending ones, but there are certainly other models that warrant discussion and would have led to an improved performance. Since most experiments were conducted in June 2024, the choice of benchmarked models reflects that. In the meantime, GPT 4o-mini has superseded GPT 3.5, LLaMa 3.3 is a continuation of LLaMa 3, Gemma 2 was released, as well as Qwen 2 after Qwen 1.5.

\section*{Acknowledgments}
We would like to thank the anonymous reviewers for their valuable suggestions. This research has been supported by the German Federal Ministry of Education and Research (BMBF) grant 01IS17049 Software Campus 2.0 (TU München).

\bibliography{custom}

\appendix
\section{Appendix}
\subsection{Additional Metrics}
The results of the experiments with gold snippets were additionally evaluated using the METEOR and cosine similarity metrics. The results are shown in Tables \ref{tab:results_metcos_bioasq} and \ref{tab:results_metcos_semqa}. For BioASQ, the results with these two metrics mostly follow the patterns observed with the original metrics from the main part, with a big jump in performance for the first snippet and then continued to increase. It is a similar case for QuoteSum, but in this dataset, the two metrics seem to peak at the top 5 snippets and then slightly drop and deteriorate when including all top 10 snippets.

\subsection{Example outputs}
Example outputs of 6 models for a question from BioASQ, together with top 3 gold snippets and ideal answer, are shown in Table \ref{tab:six_answers}.

\begin{table*}[htpb]
\small

\centering
\rowcolors{2}{gray!25}{white}
\resizebox{\textwidth}{!}{%

\begin{tabular}{r|cc|cc|cc|cc|cc|cc|cc|cc}

\rowcolor{gray!40}
\textbf{\#} & \multicolumn{2}{c}{\textbf{GPT 3.5}} & \multicolumn{2}{c}{\textbf{GPT 4o}} & 
\multicolumn{2}{c}{\textbf{LLaMa 70B}} & \multicolumn{2}{c}{\textbf{Mixtral}} & \multicolumn{2}{c}{\textbf{Gemma 7B}} & \multicolumn{2}{c}{\textbf{LLaMa 8B}} & \multicolumn{2}{c}{\textbf{Mistral}} & \multicolumn{2}{c}{\textbf{Qwen 7B}}   \\

\rowcolor{gray!40}
  \textbf{}    & \textbf{MET}       & \textbf{Cos}       & \textbf{MET}       & \textbf{Cos}      & \textbf{MET}       & \textbf{Cos}      &\textbf{MET}       & \textbf{Cos}   & \textbf{MET}       & \textbf{Cos}      & \textbf{MET}       & \textbf{Cos} & \textbf{MET}       & \textbf{Cos} & \textbf{MET}       & \textbf{Cos}      \\
\hline
\textbf{0} & 19.6 & 74.5 & 21.3 & 75.1 & 21.9 & 75.0 & 25.8 & 75.3 & 20.6 & 66.3 & 25.3 & 72.9 & 24.6 & 73.0 & 22.9 & 70.7 \\
\textbf{1} & 21.2 & 76.1 & 21.3 & 76.4 & 22.9 & 76.4 & 30.7 & 80.2 & 23.7 & 70.5 & 25.8 & 76.6 & 29.6 & 79.6 & 26.7 & 79.0 \\
\textbf{3} & 24.5 & 79.3 & 24.6 & 79.2 & 26.7 & 79.4 & 37.9 & 82.9 & 28.5 & 74.0 & 27.2 & 78.8 & 35.2 & 82.1 & 32.5 & 81.8 \\
\textbf{5} & 25.7 & 80.1 & 25.5 & 80.0 & 27.8 & 79.8 & 40.4 & 84.0 & 30.6 & 75.2 & 30.0 & 79.9 & 37.7 & 83.1 & 35.1 & 83.0 \\
\textbf{10} & 26.5 & 80.6 & 26.6 & 80.6 & 28.7 & 80.3 & 42.8 & 84.8 & 32.3 & 76.1 & 31.1 & 80.3 & 39.8 & 84.1 & 37.8 & 84.2 \\

\end{tabular}
}

\caption{\label{tab:results_metcos_bioasq} Results of final QA performance on BioASQ for different number of \textbf{gold context snippets} using the four big LLMs and four small LLMs. The results are measured with Meteor (MET) and average cosine similarity of text embeddings using \textit{all-mpnet-base-v2}.}

\end{table*}

\begin{table*}[htpb]
\small

\centering
\rowcolors{2}{gray!25}{white}
\resizebox{\textwidth}{!}{%

\begin{tabular}{r|cc|cc|cc|cc|cc|cc|cc|cc}

\rowcolor{gray!40}
\textbf{\#} & \multicolumn{2}{c}{\textbf{GPT 3.5}} & \multicolumn{2}{c}{\textbf{GPT 4o}} & 
\multicolumn{2}{c}{\textbf{LLaMa 70B}} & \multicolumn{2}{c}{\textbf{Mixtral}} & \multicolumn{2}{c}{\textbf{Gemma 7B}} & \multicolumn{2}{c}{\textbf{LLaMa 8B}} & \multicolumn{2}{c}{\textbf{Mistral}} & \multicolumn{2}{c}{\textbf{Qwen 7B}}   \\

\rowcolor{gray!40}
  \textbf{}    & \textbf{MET}       & \textbf{Cos}       & \textbf{MET}       & \textbf{Cos}      & \textbf{MET}       & \textbf{Cos}      &\textbf{MET}       & \textbf{Cos}   & \textbf{MET}       & \textbf{Cos}      & \textbf{MET}       & \textbf{Cos} & \textbf{MET}       & \textbf{Cos} & \textbf{MET}       & \textbf{Cos}      \\
\hline
\textbf{0} & 23.6 & 73.0 & 23.4 & 73.0 & 23.2 & 72.3 & 26.0 & 72.6 & 6.3 & 29.3 & 21.8 & 70.1 & 25.4 & 72.2 & 24.6 & 72.9 \\ 
\textbf{1} & 25.8 & 76.1 & 25.9 & 76.2 & 24.1 & 72.5 & 31.1 & 76.6 & 9.8 & 36.2 & 23.5 & 71.1 & 28.0 & 76.6 & 27.4 & 76.7 \\ 
\textbf{3} & 30.4 & 78.1 & 30.6 & 78.2 & 33.0 & 77.6 & 36.3 & 78.7 & 17.1 & 56.9 & 29.1 & 74.9 & 34.6 & 79.0 & 32.1 & 78.6 \\ 
\textbf{5} & 31.9 & 78.5 & 31.5 & 78.5 & 34.6 & 77.5 & 36.8 & 78.8 & 17.3 & 56.1 & 30.3 & 75.3 & 34.8 & 78.9 & 32.7 & 79.0 \\
\textbf{10} & 31.5 & 78.3 & 31.5 & 78.4 & 34.5 & 77.5 & 36.8 & 79.0 & 17.2 & 55.9 & 30.5 & 75.5 & 34.8 & 78.9 & 32.8 & 79.0 \\

\end{tabular}
}

\caption{\label{tab:results_metcos_semqa} Results of final QA performance on QuoteSum for different number of \textbf{gold context snippets} using the four big LLMs and four small LLMs. The results are measured with Meteor (MET) and average cosine similarity of text embeddings using \textit{all-mpnet-base-v2}.}

\end{table*}

\begin{table*}[htpb]
\centering
\small
\begin{tabular}{p{22mm}p{129mm}}
\hline
\textbf{Question} &  \small 
What is the role of nimotuzumab in treatment of pontine glioma?\\[0.7ex] \midrule

\textbf{Top 3 snippets} &  
(1) We found that nimotuzumab with or without radiotherapy, chemotherapy in newly diagnosed or recurrent HGG, such as glioblastoma multiforme (GBM), anaplastic astrocytomas (AA), and diffuse intrinsic pontine glioma (DIPG), might improve the response rate or the survival time. (2) In conclusion, nimotuzumab is a very well-tolerated drug with acceptable toxicity, and it may have promising value in the combination treatment. (3) Combining craniospinal irradiation (CSI) with concurrent temozolomide and nimotuzumab therapy may slightly improve tumor control and overall survival.
\\[0.8ex] \midrule

\textbf{Ideal Answer}  &  Nimotuzumab (an anti-EGFR monoclonal antibody) is being used for treatment of pontine gliomas. Nimotuzumab is a very well-tolerated drug with acceptable toxicity, and it may have promising value in the combination treatment. Clinical trials evaluating efficacy of nimotuzumab  are ongoing.  \\[0.5ex]

\hline \hline

\textbf{GPT-4} & 
The role of nimotuzumab in the treatment of pontine glioma is to potentially improve the response rate or survival time when used in combination with radiotherapy and chemotherapy.
\\[0.7ex]

\textbf{Mixtral} & 
The context suggests that nimotuzumab, a well-tolerated drug with acceptable toxicity, could potentially improve the response rate or survival time when used in combination with radiotherapy and/or chemotherapy for treating pontine glioma (DIPG) and other high-grade gliomas like glioblastoma multiforme (GBM) and anaplastic astrocytomas (AA).
\\[0.7ex]

\textbf{LLaMa 3 (70B)} & 
Nimotuzumab may improve the response rate or survival time in the treatment of pontine glioma.
\\[0.8ex] 

\textbf{LLaMa 3 (8B)} & 

 According to the context, the role of nimotuzumab in the treatment of pontine glioma is to potentially improve the response rate or survival time, either as a standalone treatment or in combination with radiotherapy and chemotherapy.
\\[2ex] 

\textbf{Mistral} & 
 Nimotuzumab may potentially improve the response rate or survival time when used in combination with radiotherapy, chemotherapy for newly diagnosed or recurrent High-Grade Gliomas (HGG), including Diffuse Intrinsic Pontine Glioma (DIPG).
\\[2ex] 

\textbf{Qwen (7B)} & 

 The role of nimotuzumab in the treatment of pontine glioma, as mentioned in the context, is that it may improve the response rate or survival time when used in combination with radiotherapy and chemotherapy for newly diagnosed or recurrent glioblastoma multiforme (GBM), anaplastic astrocytomas (AA), and diffuse intrinsic pontine glioma (DIPG). It is described as a well-tolerated drug with acceptable toxicity, suggesting it may have potential as a part of a combination therapy.
\\[0.8ex] 
    \hline

\end{tabular}
\caption{Example of a question from BioASQ with its top 3 gold snippets and ideal answer. Below are responses generated by different LLMs, demonstrating the differences in addressing the question based on the given context. }
\label{tab:six_answers}
\end{table*}

\end{document}